\documentclass[10pt,twocolumn,letterpaper]{article}

\usepackage[pagenumbers]{cvpr} 

\definecolor{darkred}{rgb}{0.7,0.1,0.1}
\definecolor{darkgreen}{rgb}{0.1,0.6,0.1}
\definecolor{green}{rgb}{0, 0.5, 0}
\definecolor{orange}{rgb}{0.8, 0.6, 0.2}
\definecolor{red}{rgb}{1.0, 0.0, 0.0}
\definecolor{teal}{rgb}{0.0, 0.4, 0.4}
\definecolor{purple}{rgb}{0.65,0.0,0.65}
\definecolor{saffron}{rgb}{0.95,0.75,0.2}
\definecolor{turquoise}{rgb}{0.0,0.5,0.5}
\definecolor{black}{rgb}{0.0, 0.0, 0.0}
\definecolor{gray}{rgb}{0.5, 0.5, 0.5}

\definecolor{cvprblue}{rgb}{0.21,0.49,0.74}
\definecolor{cvprdarkblue}{HTML}{2b75a4}

\usepackage[pagebackref,breaklinks,colorlinks,allcolors=cvprblue]{hyperref}
\usepackage{graphicx}
\usepackage{amsmath}
\usepackage{amssymb}
\usepackage{booktabs}
\usepackage{algorithm}
\usepackage{algorithmic}
\usepackage{booktabs}
\usepackage{multirow}
\usepackage{amssymb}
\usepackage{pifont}
\usepackage[table]{xcolor}
\newcommand{\cmark}{\ding{51}}
\newcommand{\xmark}{\ding{55}}

\usepackage[pagebackref,breaklinks,colorlinks]{hyperref}
\usepackage{xcolor}

\def\paperID{48} 
\def\confName{CVPR}
\def\confYear{2026}
\definecolor{cvprblue}{rgb}{0.71,0.59,0.94}
\title{\textcolor{cvprdarkblue}{FiGO}: \textcolor{cvprdarkblue}{Fi}ne-\textcolor{cvprdarkblue}{G}rained \textcolor{cvprdarkblue}{O}bject Counting without Annotations}

\author{
Adriano D'Alessandro \quad Ali Mahdavi-Amiri \quad Ghassan Hamarneh\\
Simon Fraser University\\
{\tt\small \{acdaless, amahdavi, hamarneh\}@sfu.ca}
}
\begin{document}
\maketitle
\begin{abstract}
Class-agnostic counting (CAC) methods reduce annotation costs by letting users define what to count at test-time through text or visual exemplars. However, current open-vocabulary approaches work well for broad categories but fail when fine-grained category distinctions are needed, such as telling apart waterfowl species or pepper cultivars. We present FiGO, a new annotation-free method that adapts existing counting models to fine-grained categories using only the category name. Our approach uses a text-to-image diffusion model to create synthetic examples and a joint positive/hard-negative loss to learn a compact concept embedding that conditions a specialization module to convert outputs from any frozen counter into accurate, fine-grained estimates. To evaluate fine-grained counting, we introduce \textsc{LOOKALIKES}, a dataset of 37 subcategories across 14 parent categories with many visually similar objects per image. Our method substantially outperforms strong open-vocabulary baselines, moving counting systems from “count all the peppers” to “count only the habaneros.” 

\textbf{data} --- \href{https://dalessandro.dev/datasets/lookalikes/}{dalessandro.dev/datasets/lookalikes/}
\end{abstract}
\section{Introduction}
\label{sec:intro}

\begin{figure}[t]
  \centering
\includegraphics[width=\linewidth]{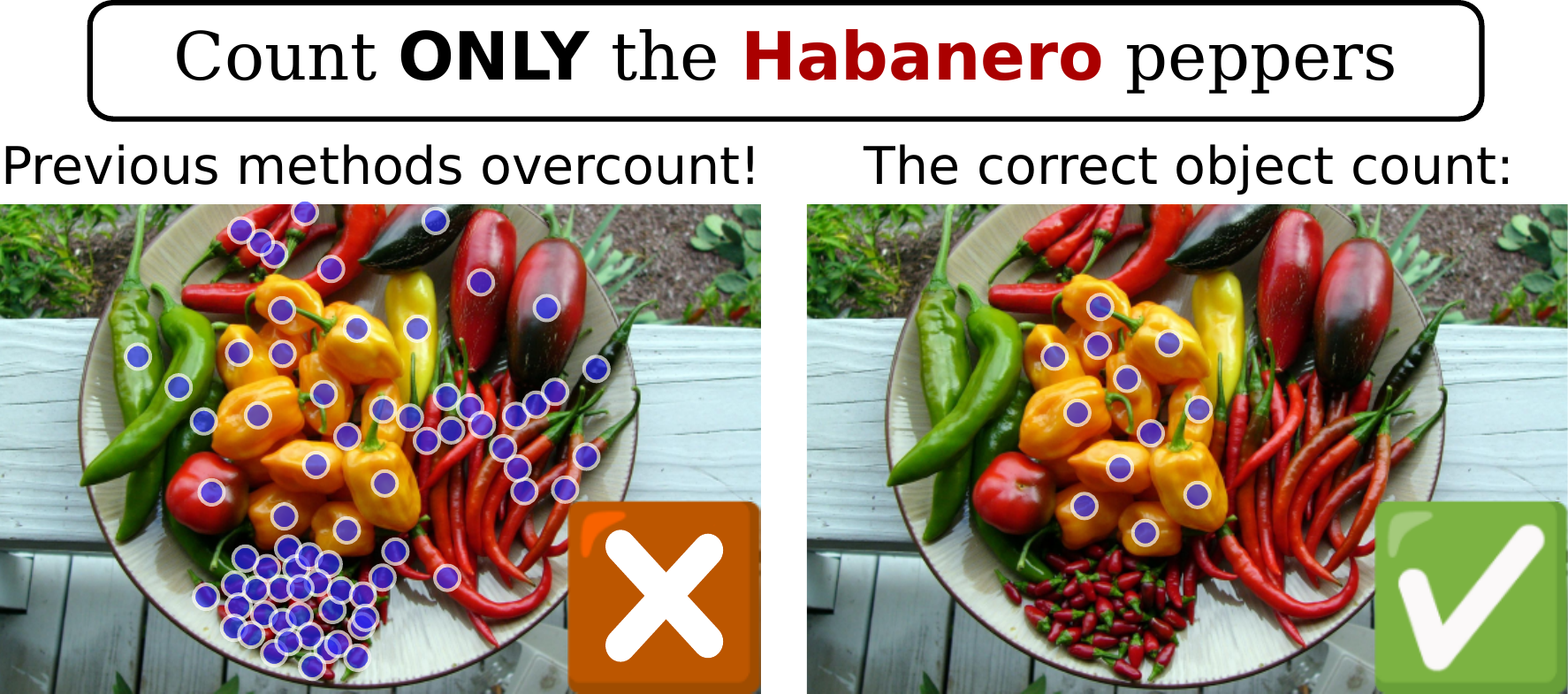}
    \caption{Overcounting is a common failure mode for open-world counting methods when presented visually similar objects.} 
    \label{fig:sneakpeak}
\end{figure}

Object counting is an important task in computer vision, supporting applications from crowd analysis to traffic monitoring. However, accurate counting typically requires manual annotations, which are expensive and time-consuming to collect. Class-agnostic counting (CAC) methods significantly reduce annotation costs by letting users define categories at inference time using text or images. However, although their recent open-vocabulary extensions~\cite{AminiNaieni24} are effective for broad object types, they do not perform as well when visual differences are subtle, such as counting only one bird species among many or identifying a single pepper cultivar in a mix (see Fig.~\ref{fig:sneakpeak}). This happens partly because models trained on limited categories do not generalize well beyond their training distribution, and text-based conditioning struggles to represent fine-grained visual concepts and category names. Prior work shows that even broad transfer is difficult: models trained on FSC-147 fail to adapt to FSCD-LVIS~\cite{chen2025single}, suggesting that limited category coverage during training leaves them underspecified.

\begin{figure}[t]
  \centering
   \includegraphics[width=\linewidth]{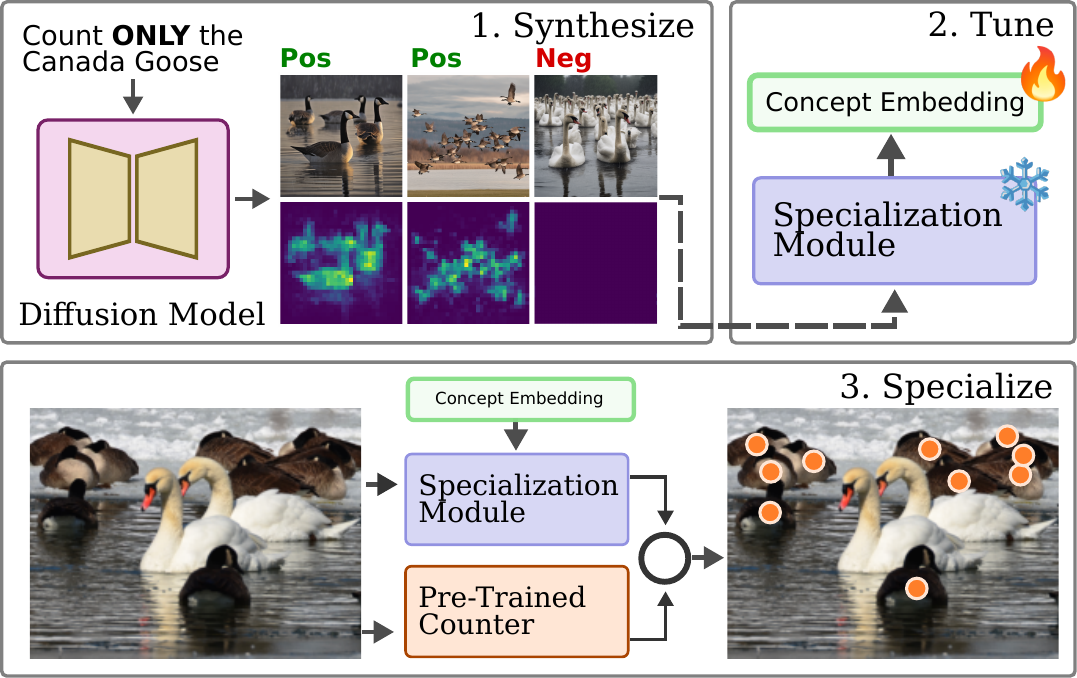}
    \caption{Given \textit{only} a text prompt, our method adapts counting models to novel fine-grained categories at test-time.}
    \label{fig:paper_overview}
\end{figure}

A straightforward response is to manually annotate examples for fine-grained categories, but this is prohibitively expensive and still does not guarantee generalization to unseen categories at test-time. Instead, we propose an annotation-free approach that preserves the efficiency of CAC while enabling reliable specialization to fine-grained categories. We achieve this by leveraging foundation text-to-image diffusion models for producing synthetic data to fine-tune concept embeddings for each new category, rather than relying only on fixed conditioning (Fig.~\ref{fig:paper_overview}).

Our approach is guided by two key insights. First, when given a sufficiently broad prompt, open-vocabulary counters tend to achieve high recall but low precision in fine-grained scenes. Thus, providing a broad prompt and then identifying these false positives can substantially improve accuracy. Second, text-to-image diffusion models can synthesize realistic fine-grained images, offering a rich source of visual supervision for learning to recognize and remove such distractors. Inspired by advances in prompt tuning for vision–language models~\cite{jia2022vpt, lester-etal-2021-power, zhou2022cocoop}, our method adapts a pre-existing counting model to novel categories by optimizing a compact concept embedding that distinguishes the target object from visually similar distractors.

Given only a category name (e.g., \textit{California Poppy}), our approach uses a text-to-image diffusion model to synthesize images and extract the attention features from between the prompt and image patches during generation. These attention maps are converted into coarse pseudo-segmentation masks that localize the named category (see: synthesis step in Fig.~\ref{fig:paper_overview}). To specialize the counter, we modify a frozen open-vocabulary segmentation model by conditioning it on concept embeddings. This module is rapidly adapted to a new category by prompt-tuning the embedding on synthetic image-mask pairs (see the tuning step in Fig.~\ref{fig:paper_overview}). For accurate distinction, we further incorporate hard negative supervision by synthesizing and using visually similar distractors as the basis for a negative concept loss. At inference time, the module is applied to the outputs of a frozen class-agnostic counter to suppress these incorrect objects, transforming overcounts into clean, category-specific estimates. Overall, our system accurately adapts to a novel fine-grained category on the fly without requiring any real images, manual annotations, or retraining of the counting model\textemdash just the category name.

Since prior class-agnostic object counting benchmarks, such as FSC147, focus on single broad categories (e.g., Boxes vs. Apples) and lack the highly specific fine-grained distinctions (e.g., Male Mallard Ducks vs. Female Mallard Ducks) necessary to test our method, they are insufficient for evaluating fine-grained counting. To address this gap, we introduce \textsc{LOOKALIKES}, a novel and challenging dataset that focuses entirely on related objects within a taxonomy. It spans 37 subcategories across 14 parent categories, and includes images containing multiple subcategories, reflecting the real-world challenge of delineating highly similar objects. Our method substantially outperforms strong open-vocabulary baselines on this new benchmark, thus addressing a critical gap in the counting pipeline: moving systems from “count all the peppers” to “count only the habaneros.” In summary, our contributions are:

\begin{itemize} 
    \item We propose FiGO, a novel annotation-free counting method that adapts to new fine-grained object categories at test-time using synthetic data.
    \item Our method serves as a plug-in module that transforms open-vocabulary counters into fine-grained counters by learning category-specific concept embeddings. 
    \item We also introduce \textsc{Lookalikes}, a new dataset designed to evaluate fine-grained counting performance and generalization across 37 subcategories.
\end{itemize}

\begin{figure}[b]
  \centering
   \includegraphics[width=\linewidth]{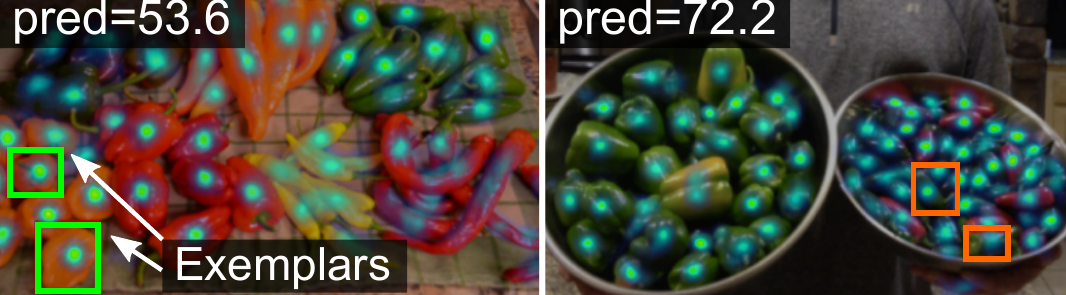}
    \caption{DAVE~\cite{Pelhan_2024_CVPR} uses a filtering strategy to reduce false positives but fails on fine-grained categories, leading to overcounting. Boxes represent visual exemplars selected from each image.}
    \label{fig:dave_compare}
\end{figure}
\section{Related Work}
\label{sec:RW}
\begin{figure*}[t]
  \centering
   \includegraphics[width=\textwidth]{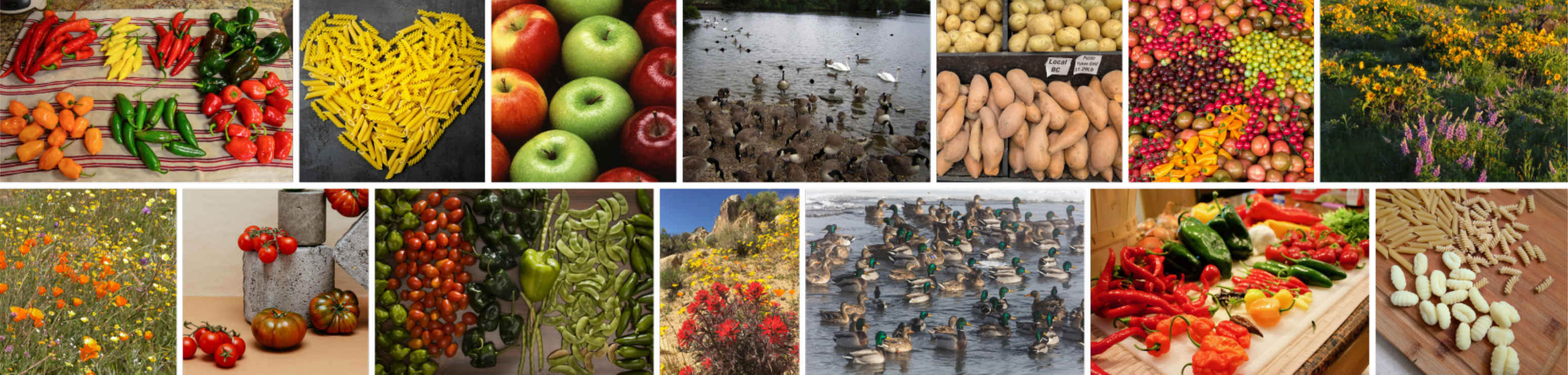}
    \caption{Sample images from \textsc{Lookalikes}, our fine-grained counting dataset. The dataset includes crowded, diverse images with multiple distinct object subcategories in each scene, highlighting the variety and complexity of the data.}
    \label{fig:dataset_samples}
\end{figure*}
\noindent\textbf{Counting with Limited Data.} Class-agnostic counting methods were introduced with the FSC147 dataset~\cite{m_Ranjane_talCVPR21}, where categories are specified at inference time. Few-shot methods use visual exemplars to define which objects to count~\cite{liu2022countr,Dukic_2023_ICCV,Pelhan_2024_CVPR,countingdetr2022}. Zero-shot counting~\cite{va_count_zhu_2024,zhizhong2024point,xu2023zero}, also known as text-shot or open-vocabulary, extended this paradigm by defining classes through text-based prompts, removing the need for visual exemplars and enabling flexible, text-driven counting. Additionally, unsupervised and weakly supervised methods~\cite{dalessandro10229913,afreeca_dalessandro,Liang_2023_CVPR,sam2019almost} have been explored for single-category object counting.

Similar to our method, DAVE~\cite{Pelhan_2024_CVPR} introduced a strategy for reducing false positives. DAVE relies on spectral clustering, which struggles with fine-grained counting (See: Fig~\ref{fig:dave_compare}), despite its filtering mechanism. Our approach learns category-specific specialization using synthetic data.

\noindent\textbf{Learning from Synthetic Images.} Synthetic image data generated by latent diffusion models is increasingly explored as a solution for real-world challenges~\cite{fake2real,he2023is, shipard2023diversity}. Closest to our approach, AFreeCA~\cite{afreeca_dalessandro} uses weak signals from synthetic images for single-category, unsupervised object counting. In contrast, our work extends this concept to multi-class fine-grained categories, using synthetic data to adapt pre-trained networks.

\noindent\textbf{Fine-Grained Image Analysis.}  Taxonomy-based analysis, as seen in datasets like CUB-200~\cite{WahCUB_200_2011} and Oxford Flowers~\cite{nilsback2008automated}, categorizes images based on taxonomies with subtle inter-category differences. Our work extends taxonomy-based analysis to annotation-free object counting.

Attribute-based analysis focuses on identifying objects with \textit{fine-grained attributes} like color or location~\cite{liu2023gres, liu2023grounding, xie2024described}. Similar to our work, the REC-8K dataset~\cite{Dai_2024_CVPR} tackles referred-expression counting, allowing for distinctions between simple categories and attributes, such as color (e.g., \textit{blue car} vs. \textit{red car}). However, REC-8K’s training and test sets share significant category overlap (75\%), focusing primarily on capturing novel attributes for established categories. In contrast, our work aims to extend to novel \textit{fine-grained categories} without requiring any annotations, addressing a previously unexplored challenge.

\noindent\textbf{Prompt Tuning.} Prompt tuning is a parameter-efficient strategy that adapts large, frozen models by optimizing a small set of learned input tokens or embeddings~\cite{jia2022vpt, zhou2022cocoop}. The authors in \cite{lester-etal-2021-power} introduced this approach for language models, showing that a small set of continuous vectors can steer model behavior without updating model weights. We extend this strategy to fine-grained dense counting.

\section{\textsc{Lookalikes} Dataset}
\begin{table}[b]
\centering
\begin{tabular}{lcccccc}
\toprule
\textbf{\# Subcategories} & \textbf{1} & \textbf{2} & \textbf{3} & \textbf{4} & \textbf{5} & \textbf{6} \\
\midrule
\textbf{\# Images} & 186  & 667  & 176  & 58 & 29 & 4 \\
\bottomrule
\end{tabular}
\caption{Distribution of subcategories per image. Most images contain 2 or more subcategories.}
\label{tab:perimage}
\end{table}

To evaluate the ability of counting models to distinguish between visually similar object categories, we introduce the \textsc{Lookalikes} dataset, a novel taxonomy-based benchmark combining fine-grained recognition with multi-class counting. Unlike broad-category counting datasets, which identify all objects of a general type (e.g., “birds” or “peppers”), our dataset focuses on distinguishing between sibling subcategories within the taxonomy, such as apple cultivars or waterfowl species. 

The \textsc{Lookalikes} dataset consists of 14 parent categories, designed to test counting models on differentiating similar objects in densely populated scenes (averaging 36.4 objects per image with a maximum of 656). Each image contains 1 to 6 sibling categories, with 83\% of images having 2 or more siblings present within the image (see: Tab~\ref{tab:perimage}). The dataset also includes images with additional, infrequently occurring subcategories labeled as “other”. These less common subcategories are grouped together and treated as a single class.

Overall, \textsc{Lookalikes} addresses a critical gap and serves as a benchmark for evaluating fine-grained object counting in high-density images with multiple subcategories from the same taxonomy, challenging models to distinguish inter-class variations within densely populated scenes. Sample images in Fig.~\ref{fig:dataset_samples} illustrate the dataset’s complexity, making \textsc{Lookalikes} a valuable resource for assessing generalization to fine-grained object categories.

\subsection{Dataset Split and Evaluation}
For easier access, we structure \textsc{Lookalikes} as a test-set only dataset with a structure similar to the test split of the FSC147 dataset~\cite{m_Ranjane_talCVPR21}. Whereas the FSC147 test-set contains 1,190 images across 29 broad categories, \textsc{Lookalikes} contains 1,120 images across 14 parent categories and 37 subcategories. This setup evaluates models’ ability to generalize to novel fine-grained categories without real images or manual annotations for fine-grained categories.

\begin{figure*}[ht]
  \centering
    \includegraphics[width=\textwidth]{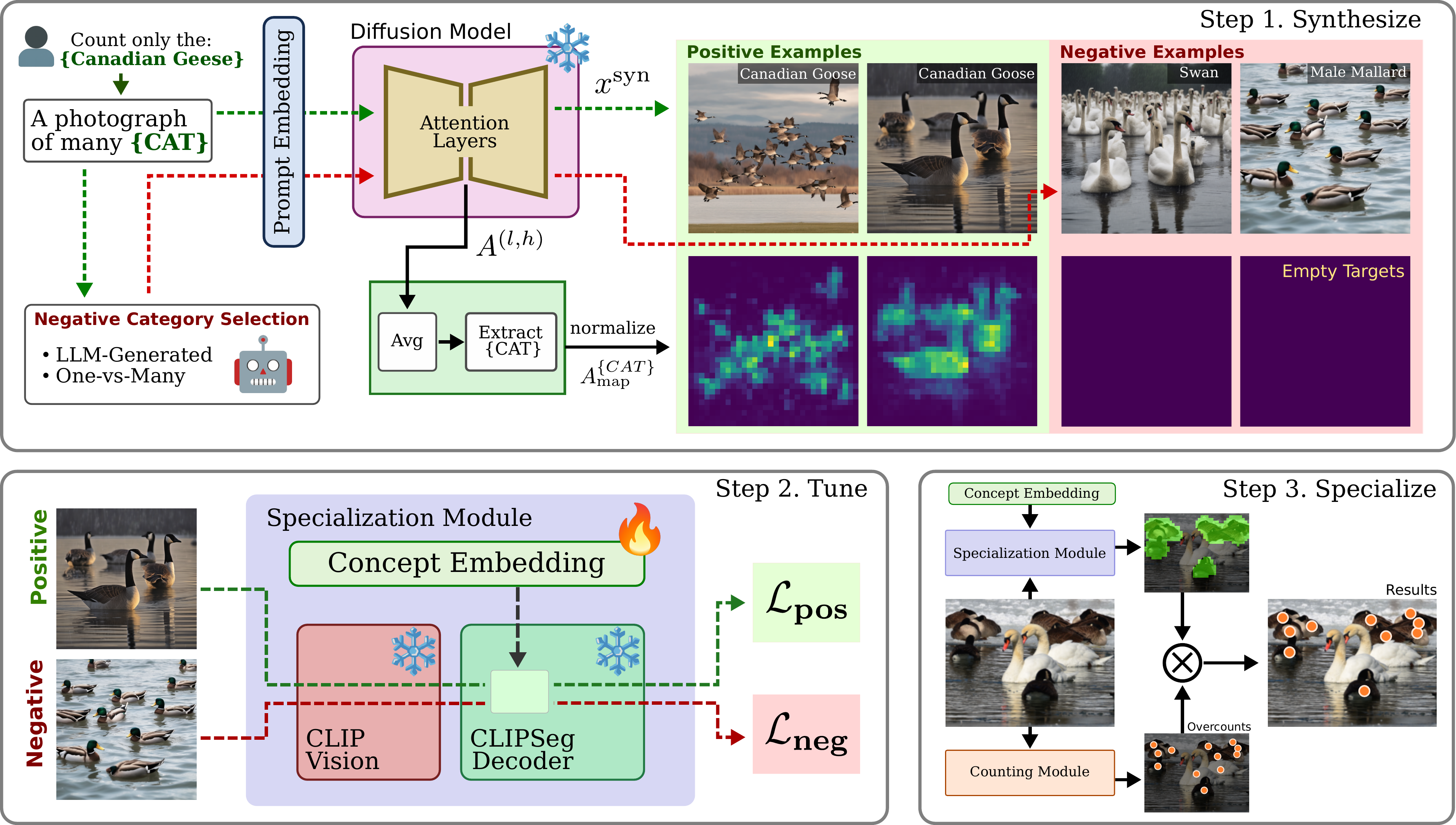}
    \caption{\textbf{Overview of our FiGO specialization pipeline for fine-grained counting.} Given a target subcategory (e.g., Canada Goose), (1) Synthesize: A diffusion model generates positive and negative examples, and category-relevant attention maps are extracted by averaging attention across layers and heads to produce dense pseudo-annotations. (2) Tune: These positive and negative synthetic pairs supervise a learnable concept embedding inside a frozen CLIPSeg model, encouraging activation primarily on the target subcategory. (3) Specialize: At inference, the tuned embedding refines the output of a frozen class-agnostic counter, yielding an accurate category-specific count.}
    \label{fig:methods}
\end{figure*}

\section{Methodology}
Here, we discuss our methodology in detail. Our method includes a prompt tuning strategy that tunes subcategory-specific concept embeddings for a specialization module using only synthetic supervision. Given only a category name, we generate synthetic images with a text-to-image diffusion model and extract semantically meaningful attention maps (Fig.~\ref{fig:methods}, Top). These maps provide sufficient guidance to tune new concept embeddings, enabling us to adapt existing counting models and improve their performance in fine-grained settings (Fig.~\ref{fig:methods}, Bottom).

The remainder of this section describes each component of our approach. We begin by outlining the generation of synthetic images and attention maps, including strategies for selecting hard negative categories. We then explain how these are used to tune concept embeddings. Finally, we describe how the resulting specialization module is applied to refine the output of a frozen class-agnostic counting model.

\subsection{Synthesize Step}\label{sec:synthesize}
Creating annotated data for fine-grained categories requires significant manual effort, making it impractical at scale. To overcome this, we use off-the-shelf foundation text-to-image diffusion models to generate synthetic images with pseudo-annotations. Synthetic images are generated as:
\begin{equation}
    x^{\text{syn}} \leftarrow \operatorname{g}(\text{prompt}),
\end{equation} where $\operatorname{g}$ is the text-to-image generator, $x^{\text{syn}}$ is the generated image, and the $\text{prompt}$ specifies the fine-grained category.


To obtain dense annotations for downstream tuning of the concept embedding, we extract the attention features between image patches and text tokens, following prior work on attention-based feature extraction~\cite{hertz2023prompttoprompt}. We first average these features across relevant heads and layers:
\begin{equation}
    A_{\text{avg}} \in \mathbb{R}^{S_{\text{text}} \times S_{\text{image}}} = \frac{1}{LH} \sum_{l=1}^{L} \sum_{h=1}^{H} A^{(l,h)},
\end{equation} where $L$ and $H$ are the number of layers and heads, $S_{\text{text}}$ is the number of tokens, and $S_{\text{image}}$ is the number of patches.

We then extract the attention between image patches and the text tokens corresponding to the fine-grained category:
\begin{equation}
A^{\{CAT\}}_{i{:}j} = A_{\text{avg}}[i:j, :],
\end{equation} where $i$ is the index of the first token in the category name and $j$ is the index of the last token. If the category name spans multiple tokens, we select the token that maximally attends to image regions. This yields a spatial attention map $A^{\{CAT\}}_{\text{map}} \in \mathbb{R}^{H \times W}$, which is reshaped to match the image patch grid and normalized to $[0,1]$:
\begin{equation}
\hat{A}^{\{CAT\}}_{\text{map}} = \operatorname{norm}(A^{\{CAT\}}_{\text{map}}).
\end{equation}

The resulting pair $(x^{\text{syn}}, \hat{A}^{\{CAT\}}_{\text{map}})$ provides a synthetic image and a dense pseudo-annotation highlighting object locations. This process automatically generates pseudo-annotated datasets for fine-grained concept specialization.

\subsection{Tuning the Specialization Module}
To adapt the model to a novel fine-grained category at test time, we tune a specialization module that suppresses visually similar distractors and preserves the target object. This is achieved by optimizing a subcategory-specific concept embedding, $z_{concept}$, using pseudo-annotated synthetic images. The embedding replaces the conditioning in a frozen open-vocabulary segmentation network. We use CLIPSeg~\cite{lueddecke22_cvpr} for this purpose due to its accessibility. Because our goal is not precise segmentation but spatially-aware category specialization, we treat the coarse attention maps from synthetic exemplars as sufficient guidance for aligning the embedding with the visual concept of interest. Further, we use positive and negative supervision to guide the embedding toward precise activation on the target category and away from visually similar but incorrect regions.

\subsubsection{Positive Concept Supervision}
To guide the concept embedding toward the target category using $\hat{A}^{\{CAT\}}_{\text{map}}$, we apply a fixed threshold to obtain a binary mask, $A^{\{CAT\}}_{\text{bin}}$, that localizes the category.

Each synthetic image, $x^{\texttt{syn}}$, is then preprocessed by the CLIP vision encoder to produce image features $F^{\texttt{syn}}$. We then pass both $z_{concept}$ and $F^{\texttt{syn}}$ to the CLIPSeg decoder, which outputs a predicted binary mask $\hat{A}_{\text{pred}} \in [0, 1]^{H \times W}$. The concept embedding $z_{concept}$ is optimized using binary cross-entropy calculated as follows:

\begin{equation}
\mathcal{L}_{\text{pos}} = \text{BCE}(\hat{A}_{\text{pred}}, A_{\text{bin}}^{\{CAT\}})
\end{equation}

This encourages the concept embedding to activate in the correct spatial regions associated with the target category.

\begin{table*}[t]
\centering
\caption{\textbf{Evaluation of counting and detection methods on the \textsc{Lookalikes} dataset.} FiGO specialization significantly improves prior counting networks on fine-grained categories. Methods marked with ${\dagger}$ use a \textsc{Broad} prompt, and others use a \textsc{Fine}-grained prompt. Bold values indicate, for each base network, the better of its two variants (with FiGO vs. without). Underlined values denote the best overall performance. The Multi-class and Real data columns indicate training on multi-category images or real images, respectively. }
\setlength{\tabcolsep}{1mm}  
\begin{tabular}{lccccccccc}
\toprule
\textbf{Method} & \textbf{Text Encoder} & \textbf{Category Handling} & \textbf{Scope} & \textbf{Multi-Class} & \textbf{Real Data} & 
\boldmath{$\text{MAE}$} &  
\boldmath{$\text{RMSE}$} & 
\boldmath{$\text{MRAE}$} \\
\midrule
OwlV2~\cite{minderer2024scaling} 
    & CLIP & Text &  Detect & \cmark & \cmark & 23.24 & 36.31 & 1.05  \\
Gr.Dino~\cite{liu2023grounding} 
    & BERT & Text &  Detect & \cmark & \cmark & 25.89 & 39.63 & 0.90 \\
    
\midrule
PseCo~\cite{zhizhong2024point} 
    & CLIP & Text &  Count & \xmark & \cmark & 42.88 & 56.67 & 5.92 \\
\rowcolor{gray!20}  ~~~\textit{FiGO} \textcolor{darkgray}{+ PseCo}$^{\dagger}$ 
    &-&Prompt Tune& Count & \xmark & \xmark &  \textbf{19.57} & \textbf{30.86} & \textbf{0.79} \\
DAVE~\cite{Pelhan_2024_CVPR} 
    & CLIP & Text &  Count & \xmark & \cmark & 53.10 & 64.44 & 6.52  \\
\rowcolor{gray!20}  ~~~\textit{FiGO} \textcolor{darkgray}{+ DAVE}$^{\dagger}$     
    &-& Prompt Tune & Count & \xmark & \xmark &  \textbf{21.19} &  \textbf{31.53} & \textbf{0.86}  \\
CountGD~\cite{AminiNaieni24} 
    & BERT & Text &  Count & \xmark & \cmark &  30.17 & 45.01 & 3.35  \\
\rowcolor{gray!20}  ~~~\textit{FiGO} \textcolor{darkgray}{+ CountGD}$^{\dagger}$     
    &-& Prompt Tune & Count & \xmark & \xmark &  \underline{\textbf{13.31}} & \underline{\textbf{23.36}} & \underline{\textbf{0.58}} \\
GroundingREC~\cite{Dai_2024_CVPR}
    & BERT & Text &  Count & \cmark & \cmark &  18.51 &  30.39 & \textbf{0.81}  \\
\rowcolor{gray!20} ~~~\textit{FiGO} \textcolor{darkgray}{\small{+ GroundingREC}}$^{\dagger}$     
    & - & Prompt Tune & Count & \xmark & \xmark &  \textbf{16.15} &\textbf{26.99} & 0.86  \\
\bottomrule
\end{tabular} \\

\label{tab:main_results}
\end{table*}
\subsubsection{Negative Concept Supervision}\label{sec:negative_concept}
Images containing negative categories help refine the concept embedding by suppressing activations for visually similar but incorrect classes. Selecting informative negative categories requires careful consideration. We use LLM-generated negatives for nearly all experiments but explore the following two strategies for generating negatives:

\vspace{4pt}
\textit{LLM-Generated.} We use a large language model to produce lists of visually similar sibling categories for each target class. These categories are used to synthesize hard negatives that provide targeted distractors for concept refinement. This is the default approach in most experiments.

\vspace{4pt}
\textit{Fine-vs-Broad.} In a one-vs-many setup, we synthesize images from a broader parent category (e.g., for the target Canada Goose, we use the prompt ``Waterfowl''). This yields negatives that include visually similar examples.

\vspace{4pt}
For each negative category, we synthesize images as described in Section~\ref{sec:synthesize} and assign an empty attention map:
\begin{equation}
A^{\neg\text{\{CAT\}}}_{\text{bin}} = 0.
\end{equation}
These examples discourage responses on distractor regions. The negative concept loss is defined as:
\begin{equation}
\mathcal{L}_{\text{neg}} = \operatorname{BCE}(\hat{A}_{\text{pred}}, A^{\neg\text{\{CAT\}}}_{\text{bin}}).
\end{equation}
Finally, the total concept tuning loss combines positive and negative supervision:
\begin{equation}
\mathcal{L}_{\text{concept}} = \mathcal{L}_{\text{pos}} + \mathcal{L}_{\text{neg}}.
\end{equation}

\subsection{Specialization}
After tuning the category-specific concept embedding, we apply the specialization module to refine the raw predictions of the class-agnostic counting model. Given a real input image $x$ and the tuned concept embedding $z_{\text{concept}}$, we pass both through the specialization module to produce a spatial mask $\hat{A}_{\text{pred}}$ that represents the predicted relevance of each pixel to the target category. We then pass the image through the class-agnostic counter to produce $D_{\text{raw}}$, which may be either a density map or a point map. To suppress false positives, we re-weight the output using the predicted mask: $D_{\text{special}} = D_{\text{raw}} \odot \hat{A}_{\text{pred}}$.
This step preserves true positives while ignoring distractor regions, resulting in a category-specific output $D_{\text{special}}$ that reflects only the target class.

\section{Experiments}
\setlength{\tabcolsep}{1mm}
\begin{table}[t]
\centering
\caption{\textbf{FiGO vs. Open-Vocab. Segmentation.} Comparing specialization strategies using CountGD as the base counter.}
\begin{tabular}{lccc}
\toprule
\textbf{Method} & \boldmath{$\text{MAE}$} & \boldmath{$\text{RMSE}$} &  \boldmath{$\text{MRAE}$}\\
\midrule
CLIPSeg~\cite{lueddecke22_cvpr}      & 22.20 & 35.61   & 0.86 \\
OVSeg~\cite{liang2023open}    & 25.41 & 36.62 & 0.96  \\
Ours       & \textbf{13.31} & \textbf{23.36} & \textbf{0.58}  \\
\bottomrule
\end{tabular}

\label{tab:open_comparison}
\end{table}

\begin{figure*}[t]
  \centering
\includegraphics[width=\linewidth]{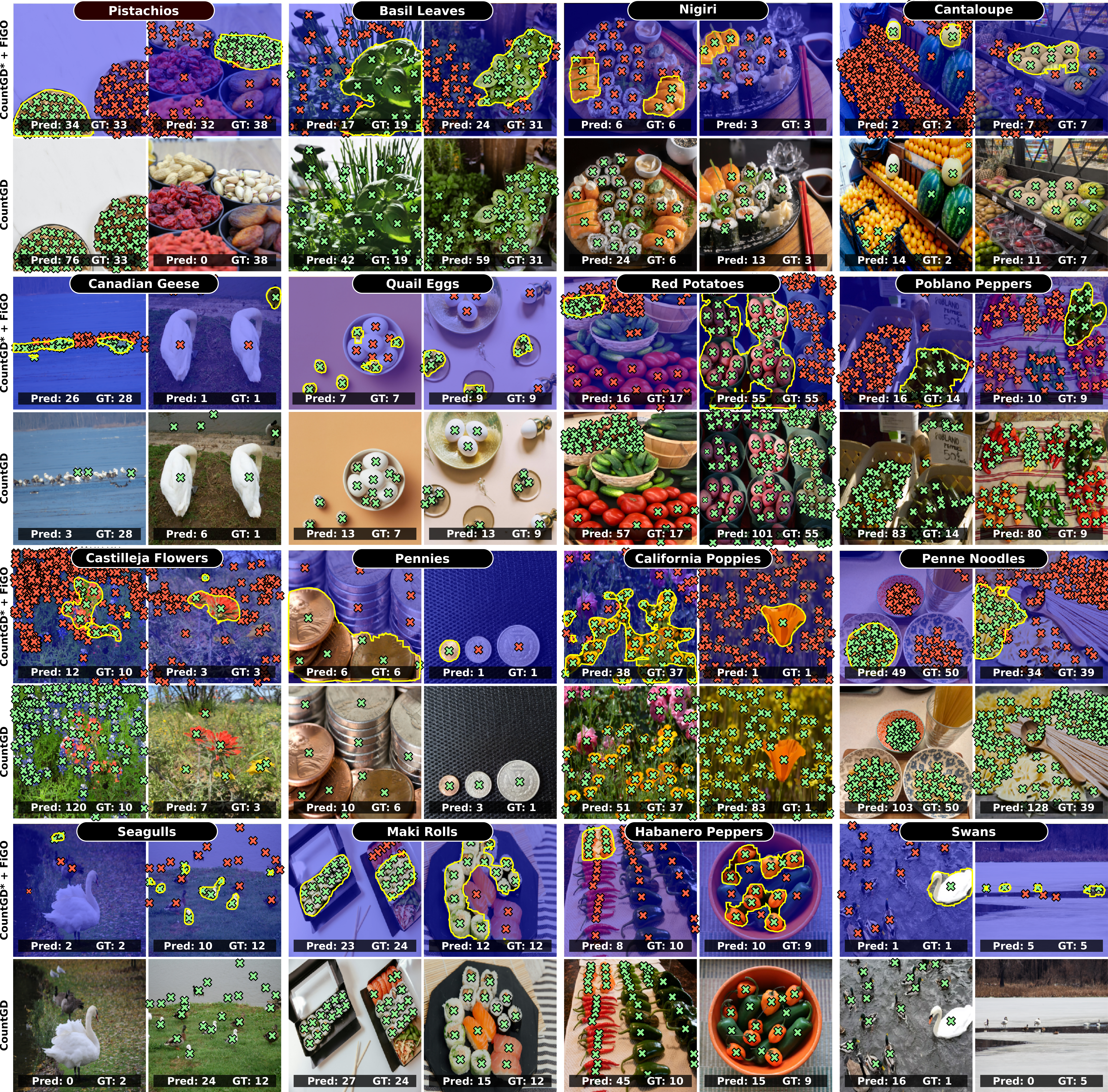}
   \caption{\textbf{Qualitative results on the \textsc{LookAlikes} dataset comparing CountGD with and without FiGO.} Each column shows two images from sixteen fine-grained categories (e.g., basil leaves, nigiri, quail eggs). The upper row in each pair shows our FiGO specialization applied to CountGD, where a broad prompt (e.g., nuts, birds, leaves) is used to deliberately encourage overcounting, and the fine-grained category name is used to tune the concept embedding that filters these predictions. Yellow masks indicate retained regions after specialization, green crosses denote retained points, and red crosses denote discarded points. The lower row shows the baseline CountGD using only the fine-grained category name as the prompt. Without FiGO, CountGD overcounts visually similar distractors (e.g., surrounding foliage near basil leaves) or undercounts categories where the fine-grained label is inconsistently recognized (e.g., pistachios). FiGO suppresses false positives while preserving relevant instances even in challenging scenarios, such as separating sweet potatoes from red potatoes or cilantro leaves from basil leaves. Our method yields accurate, category-specific counts across all categories, highlighting FiGO’s generalizability and ability to specialize across varied visual domains.}
    \label{fig:qualitative}
\end{figure*}


\paragraph{Evaluation Criteria.} We evaluate all methods using mean absolute error (MAE), root mean squared error (RMSE), and mean relative absolute error (MRAE). To account for dataset imbalance, we first compute the average error per subcategory (mean over all images containing it) and then average across subcategories:
\begin{equation}
\text{Metric} = \frac{1}{C} \sum_{c=1}^{C} \left( \frac{1}{|I_c|} \sum_{i \in I_c} \mathcal{E}(y_i, \hat{y}_i) \right),
\end{equation}
\noindent where $C$ is the number of subcategories, $I_c$ is the set of images containing subcategory $c$, $\mathcal{E}(y_i, \hat{y}_i)$ is the error metric.
\setlength{\tabcolsep}{1mm}
\begin{table}[t]
\centering
\caption{\textbf{Synthetic Data Source.} Comparison of different diffusion models as sources of synthetic data. Speed represents the generation time for an image on an A100 GPU.}
\begin{tabular}{lcccc}
\toprule
\textbf{Method} & \textbf{Speed (s/img)} & \boldmath{$\text{MAE}$} & \boldmath{$\text{RMSE}$} &  \boldmath{$\text{MRAE}$}\\
\midrule
SDXL~\cite{podell2024sdxl}   & 1.4 & 16.29 & 27.01 & 0.78  \\
SD3-\footnotesize{Med}~\cite{esser2024scaling}    & 0.9 & 16.05 & 26.20 & 0.71  \\
Flux.1-\footnotesize{Schnell}~\cite{labs2025flux1kontextflowmatching}  & \textbf{0.4} & \textbf{13.31} & \textbf{23.36} & \textbf{0.58}  \\
\bottomrule
\end{tabular}
\label{tab:generator}
\end{table}
\paragraph{Main Results.} We evaluate several strong baselines on the \textsc{LookAlikes} dataset with and without FiGO specialization. Given that FiGO only relies on a user-specified text prompt to define the target category, we focus on text-conditioned baselines, including class-agnostic counting methods and open-vocabulary detection models. Together, these baselines represent a broad range of strategies for handling novel or fine-grained categories.

For CAC methods, we compare against PSeCo~\cite{zhizhong2024point}, DAVE~\cite{Pelhan_2024_CVPR}, CountGD~\cite{AminiNaieni24}, and GroundingREC~\cite{Dai_2024_CVPR}. For open-vocabulary detection, we include OWL-V2~\cite{minderer2024scaling} and GroundingDINO~\cite{liu2023grounding}. We also indicate whether each method was trained on real data and in a multi-class setting.

As shown in Table~\ref{tab:main_results}, FiGO improves performance across all object counting models on the \textsc{LookAlikes} dataset, highlighting the effectiveness of our prompt tuning strategy for fine-grained specialization. The overall best-performing model was CountGD with FiGO, achieving a MAE of 13.31. The strongest baseline without FiGO was GroundingREC, with a MAE of 18.51. The performance of GroundingREC likely reflects training on a large dataset with many multi-class scenes. However, it experiences performance drops for some categories, especially when the broader category appeared in its training set. For example, it struggled with ``Chicken Eggs'' whereas ``White Eggs'' was a training category. This suggests overfitting to simpler attribute–category combinations.

\paragraph{Comparison to Open-Vocabulary Segmentation.}
In Table~\ref{tab:open_comparison}, we compare our approach to the open-world segmentation methods OVSeg~\cite{liang2023open} and CLIPSeg~\cite{lueddecke22_cvpr}. Our method consistently outperforms other methods on the \textsc{Lookalikes} dataset, underscoring the importance of our method for reducing false positives in fine-grained counting tasks. We attribute this improvement to the difficulty these open-vocabulary segmentation methods have in distinguishing between closely related subcategories, similar to those faced by class-agnostic counting models.

\paragraph{Qualitative Analysis.} Figure~\ref{fig:qualitative} presents qualitative results from the \textsc{LookAlikes} dataset, illustrating the precision of our method across diverse categories. We evaluate CountGD with and without FiGO specialization. For CountGD, we use the fine-grained category name as the text prompt. However, CountGD occasionally misses fine-grained categories. Therefore, when applying FiGO we instead use the broadest corresponding category to deliberately encourage overcounting (e.g., “Nuts” rather than “Pistachios”).

CountGD shows a range of behaviors when applied to fine-grained categories. In some cases, it undercounts subtle targets (e.g., missing “Pistachios”) or produces extra detections when visually similar distractors are present (e.g., background leaves for “Basil Leaves”). Prior work has noted that open-vocabulary models, such as GroundingDINO, can find fine-grained prompts challenging~\cite{bianchi2024devil}, and CountGD may reflect some of these characteristics. FiGO addresses this by starting from a broader concept and then specializing, which recovers fine-grained distinctions.

Our method is also robust to unseen negative categories. For instance, the first Pistachios image and the second California Poppy image in Figure~\ref{fig:qualitative} contain distractors (Peanuts and California Goldfields, respectively) that were not included in the negative set during concept tuning.

\paragraph{Synthetic Data Source.} In Table~\ref{tab:generator}, we analyze how the choice of diffusion model influences final performance. Different generators vary in their ability to capture fine-grained details and produce realistic scenes. We use CountGD as the baseline counter and evaluate three widely used text-to-image models: SDXL~\cite{podell2024sdxl}, SD3-Medium~\cite{esser2024scaling}, and Flux.1-Schnell~\cite{labs2025flux1kontextflowmatching}. We compare their generation accuracy and speed on an A100 GPU and observe that Flux.1-Schnell provides the strongest overall results. Notably, all generators yield improvements that exceed the best standalone counting method.

\paragraph{Number of Synthetic Images.}
Given that image synthesis is the primary source of computational cost in our pipeline, and this cost scales linearly with the number of generated images, we evaluate whether similar performance can be achieved with fewer synthetic examples.

\begin{figure}[t]
  \centering
\includegraphics[width=\linewidth]{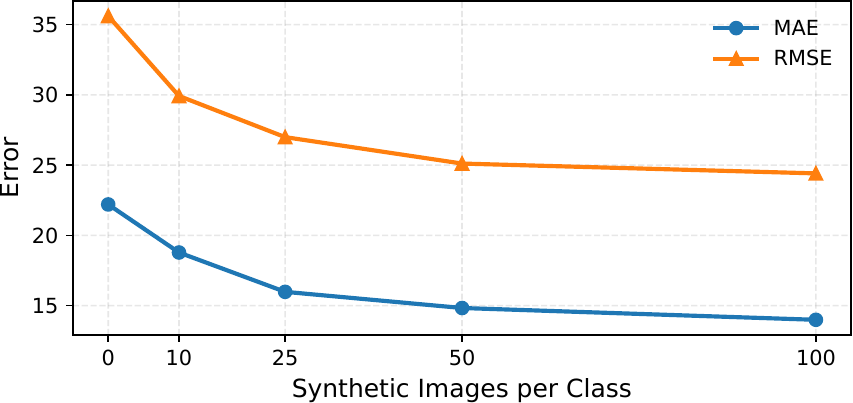}
    \caption{\textbf{Performance vs. Image Count.} FiGO performance improves as the number of synthetic images increases.}
    \label{fig:plot_image_count}
\end{figure}

In Figure~\ref{fig:plot_image_count}, we vary the number of synthetic images used to tune the concept embedding, reducing the 100 positive and 100 negative images used in our main experiments to 50, 25, 10, and 0. The zero-image setting corresponds to using an unmodified CLIPSeg model.

Even with only 50, 25, or 10 images per category, our method continues to outperform strong baselines such as CountGD by a substantial margin. These settings represent approximately 50\%, 75\%, and 90\% reductions in computational cost, respectively, with only modest performance loss. As shown in Figure~\ref{fig:plot_image_count}, performance decreases smoothly as the number of synthetic images decreases, demonstrating the robustness of our approach under constrained resource settings.

\setlength{\tabcolsep}{1mm}
\begin{table}
\centering
\caption{\textbf{Comparison of different strategies for selecting negative categories.} LLM-generated hard negatives provide stronger suppression of irrelevant objects.}
\begin{tabular}{lccc}
\toprule
\textbf{Negative Source} & \boldmath{$\text{MAE}$}  &  \boldmath{$\text{RMSE}$} & \boldmath{$\text{MRAE}$}\\
\midrule
No Negatives         & 25.68 & 35.31 & 1.56 \\
Fine-vs-Broad        & 18.91 & 31.56 & 0.81 \\
LLM-Generated        & \textbf{13.31} & \textbf{23.36} & \textbf{0.58}  \\
\bottomrule
\end{tabular}

\label{tab:ablation-negatives}
\end{table}
\setlength{\tabcolsep}{1mm}
\begin{table}[t]
\centering
\caption{\textbf{Impact of the initialization strategy for the concept embedding.} Initializing the concept embedding using the CLIP text embedding outperforms random initialization. }
\begin{tabular}{lccc}
\toprule
\textbf{Method} & \boldmath{$\text{MAE}$} & \boldmath{$\text{RMSE}$} &  \boldmath{$\text{MRAE}$}\\
\midrule
Random    & 18.20 & 30.19 & 0.76  \\
CLIP Text     & \textbf{13.31} & \textbf{23.36} & \textbf{0.58}  \\
\bottomrule
\end{tabular}

\label{tab:initialization}
\end{table}

\paragraph{Negative Category Selection.}
Table~\ref{tab:ablation-negatives} presents an ablation study comparing the negative category selection strategies introduced in the methodology section. We also evaluate a setting where negative supervision is removed entirely. This analysis shows how each strategy affects the model’s ability to suppress visually similar distractors.

Even without negative supervision, our method provides a small improvement to CountGD. The Fine-vs-Broad strategy offers reasonable performance, but the LLM-Generated negatives yield the best suppression and overall results.

\paragraph{Initialization Strategy.} In Table~\ref{tab:initialization}, we examine how the initialization of the concept embedding affects performance. We compare two choices: random initialization and initialization from the CLIP text embedding of the category. Because our specialization module builds on CLIPSeg, we expect the CLIP-based initialization to provide a stronger prior and reduce overfitting to synthetic data by beginning closer to a semantically meaningful solution. The results in Table~\ref{tab:initialization} confirm this assumption, showing that initializing from the CLIP text embedding leads to better performance.


\begin{table}[t]
    \centering
    \setlength{\tabcolsep}{4pt}

    \caption{
        \textbf{Inference-time Throughput Comparison.} We compare FiGO against commonly used components in modern counting pipelines. \textit{Params.} denotes the number of model parameters, \textit{Res.} the default input resolution for each method, \textit{GPU} the device on which inference is measured, and \textit{FPS} the achieved frames per second (higher is better). Methods annotated with (*) indicate a compiled, hardware-optimized version provided by the authors.
    }
    \begin{tabular}{lrclr}
        \toprule
        Method & Params. & Res. & GPU & FPS  \\
        \midrule
        SAM2 (Hiera-B+)* & 80.8M &$1024^2$& A100 &  43.8  \\
        SAM2 (Hiera-B+) & 80.8M &$1024^2$ & A100  &  20.8   \\
        GroundingDino & 172.2M &$800^2$ & TITAN X &  3.9 \\
        GroundingDino & 172.2M &$800^2$ & H100  &  9.8 \\
        \midrule
        FiGO (ours) & 150.7M & $356^2$  & TITAN X  & 39.9   \\
        FiGO (ours) & 150.7M & $356^2$ & H100 & 91.3  \\
        \bottomrule
    \end{tabular}

    \label{tab:inference_cost}
\end{table}
\section{Computational Cost}
This section evaluates the computational efficiency of FiGO in both inference and specialization. We report inference-time throughput relative to commonly used components in modern counting pipelines and separately measure the one-time cost of tuning a new concept embedding. This breakdown highlights that FiGO is lightweight during specialization and introduces only minimal overhead at deployment.
\paragraph{Inference Costs.} Table \ref{tab:inference_cost} reports the additional inference-time cost introduced by FiGO after specialization is complete. Given that our method only needs to be specialized once-per-category, the long-term computational burden is primarily realized during inference. Since many recent counting methods adopt SAM/SAM2 \cite{Kirillov_2023_ICCV, ravi2024sam2} as a post-processing step \cite{AminiNaieni24, hobley2023abc} or use GroundingDINO \cite{liu2023grounding} as a base module \cite{AminiNaieni24, Dai_2024_CVPR}, we benchmark FiGO against both. All models are evaluated at their default input resolutions. We measure FiGO and GroundingDINO performance on an H100 (80 GB) and a TITAN X (12 GB) to assess behavior across hardware settings, and we use prior reported SAM2 performance on an A100 (80 GB). FiGO runs at 91.3 FPS on the H100 and 39.9 FPS on the TITAN X. Relative to GroundingDINO, FiGO processes 9–10× more images per second, and compared to SAM2, FiGO achieves 2–4× higher throughput, depending on whether SAM2 is compiled for the target GPU. Our method provides substantial performance gains to existing counting models while adding minimal overhead during inference. 

\paragraph{Tuning Costs.} There are two sources of computational burden introduces when tuning a new concept embedding. Synthetic data generation forms the bulk of the computational burden for our method. Generating 100 positive and 100 negative examples on an H100 GPU using FLUX.1-schnell takes approximately 3-4 minutes including warm-up time. Within the same setting, tuning the concept embedding using the synthetic data takes 1-2 minutes. In total, the full specialization process takes only 4-6 minutes per category and is only performed once per category. Because our method does not require any real images or annotations, it remains practical for real-world applications where collecting and annotating real data is expensive and time-consuming.


\section{Implementation}
We provide the full implementation details of our specialization process, including the hyperparameters used during tuning, the augmentations applied to synthetic data, the training and model-selection procedure, the choice of attention layers for each diffusion model, and the configuration used when evaluating baseline methods on \textsc{LookAlikes}.

\paragraph{Implementation.} Synthetic images are generated using Flux.1-Schnell~\cite{labs2025flux1kontextflowmatching} at a resolution of $720 \times 720$ with 4 inference steps, producing 100 positive and 100 negative images per category (utilizing 5 LLM-generated negative categories per positive one). For specialization, we employ CLIPSeg with \texttt{CIDAS/clipseg-rd64-refined} weights. The concept embedding $z_{\text{c}}$ comprises 512 parameters initialized via the CLIP text encoder and is tuned for 50 epochs using the AdamW optimizer~\cite{loshchilov2018decoupled} with a learning rate of $5\times10^{-3}$. We reserve 20\% of the synthetic data for validation, reducing the learning rate by a factor of 0.9 if the validation loss does not improve for 10 epochs. Additionally, pseudo-attention maps are binarized using a fixed threshold of 0.1. Hyperparameters were determined using synthetic data.
\paragraph{Hyperparameters.}  We tune the concept embedding $z_{\text{c}}$ for 50 epochs using the AdamW optimizer~\cite{loshchilov2018decoupled} with a learning rate of $5\times10^{-3}$. We reserve 20\% of the synthetic data as a validation set for monitoring training. The learning rate is reduced by a factor of 0.9 if the validation loss does not improve for 10 epochs. Pseudo-attention maps generated by the diffusion model are binarized using a fixed threshold of 0.1. We use 5 LLM-generated negative categories per positive category. Hyperparameters were determined using synthetic data from several categories.

\paragraph{Augmentation.} We apply two forms of augmentation during specialization. First, we downscale each synthetic image to 50\% of its original size and then pad it back to the target resolution, increasing the diversity of object scales encountered during tuning. Second, we use CutMix~\cite{yun2019cutmix} in feature space by swapping spatial quadrants between positive and negative samples. This helps prevent the model from overfitting to single-category images and encourages robustness to mixed-scene compositions.

\paragraph{Training.} Before optimization, all synthetic images are resized to $356 \times 356$, and visual features are pre-extracted using the CLIP vision encoder to reduce computational cost during tuning.

For model selection, we monitor both the validation loss and the sharpness of the learned concept embedding, which measures its stability under small perturbations. Prior work shows that flatter minima tend to generalize better~\cite{keskar2017on}, whereas sharp minima would indicate overfitting to synthetic data and pseudo-annotations. We define sharpness as

\begin{equation}
s(z_{\text{c}})
= 
\frac{1}{K}
\sum_{k=1}^K 
\max\!\big(0,\; \mathcal{L}(z_{\text{c}} + \delta_k) - \mathcal{L}(z_{\text{c}}) \big),
\end{equation} where $\delta_k$ denotes a small random perturbation of the concept embedding and $\mathcal{L}$ is the concept loss. Sharpness estimates the worst-case increase in loss within a small neighborhood around the current parameter value. During selection, we first retain only the top 20\% of epochs with the lowest sharpness, and from this subset choose the checkpoint with the best synthetic validation performance.

\paragraph{Attention Layers.} We observed that different diffusion models exhibit varying attention behaviors across their internal blocks. For SDXL, averaging the cross-attention maps over all blocks produced stable and accurate pseudo-attention maps. In contrast, for FLUX.1, manual inspection revealed that blocks 3, 4, and 5 consistently attended most strongly to the target objects across all categories, yielding cleaner and more localized maps than averaging over all layers. Accordingly, we aggregate attention only from these blocks when using FLUX.1.

\paragraph{Evaluation Settings.} In our main results table, we evaluate several methods on the \textsc{LookAlikes} dataset. Below, we describe the exact settings used for each method.

We use the official GitHub implementations and pre-trained weights released by the respective authors for all counting and detection models. We do not train any models from scratch or re-implement any components; all methods are run using their default configurations. Each model is prompted with the corresponding fine-grained category name (e.g., ``California Poppy,'' ``Canada Goose''), except for GroundingREC. We observed that GroundingREC produced suboptimal results when prompted solely with the category name. After testing multiple prompt templates, we found the model performed best when appending the phrase ``in an image,'' such as ``California Poppy in an image.'' This is likely because GroundingREC is trained explicitly for referring expressions rather than single category names.

For GroundingDINO and OwlV2, we use their HuggingFace implementations. GroundingDINO is initialized with the \textsc{IDEA-Research/grounding-dino-tiny} checkpoint, and for OwlV2 we use \textsc{google/owlv2-base-patch16}. We use a box threshold of 0.4 and a text threshold of 0.3 for GroundingDINO, and a box threshold of 0.2 for OwlV2.

\section{Prompt Selection}
\paragraph{Negative Selection with LLMs.} We proposed an LLM-based strategy for generating hard-negative category suggestions, which are used during finetuning the concept embedding. We rely on OpenAI's GPT5 as the LLM and use the following prompt to generate the list:
\begin{quotation}
\noindent``Provide a diverse list of exactly \{COUNT\} object categories that are semantically similar to \{CATEGORY\} and are very likely to appear in the same everyday environment. The items should be familiar categories that are either visually similar or from a closely related taxonomy. Only provide the list and nothing else.''
\end{quotation}
\paragraph{Prompt Templating.} We do basic prompt templating to encourage diversity when generating synthetic images. We use the following prompt harness:

\begin{quotation}
\noindent ``A photorealistic image of \{COUNT\} \{CATEGORY\}. \{VIEW\}, \{LIGHT\}''
\end{quotation}

Where \{COUNT\} is set to one of [``many'', ``hundreds of'', ``a few'', ``exactly two''], and \{VIEW\} is set to one of [``top-down view'',``high angle'',``viewed from a distance'',``close-up'',``macro shot''], and \{LIGHT\} is set to one of [``backlit'', ``soft lighting'', ``golden hour'', ``overcast'', ``sunlight'', ``dimly lit''].
\section{Extended Dataset Details}
\paragraph{Data Collection and Annotation Process}
To construct the \textsc{LookAlikes} dataset, we collected images under Creative Commons licenses from Flickr, Unsplash, and Pexels, using targeted search queries based on each object’s subcategory name. Selection criteria focused on maintaining crowd density, visual variability, and the presence of multiple subcategories within each image, creating realistic scenarios for fine-grained counting. For annotations, we provide the object counts along with point labels applied to the approximate center of each object within an image, with each label specifying the fine-grained category.

\section{Conclusions}
We present FiGO, a simple yet powerful approach for fine-grained object counting. By leveraging synthetic data and tuning concept embeddings on the fly, our method transforms overcounts into accurate, category-specific counts---without requiring real images or manual annotations.

\noindent\textbf{Limitations.} While effective on natural images, our method relies on the capacities of diffusion models. In domains with limited coverage, such as medical imaging, its applicability may be reduced. Future work could address this by incorporating methods like Break-A-Scene~\cite{avrahami2023bas} to enable concept learning in underrepresented domains.

{
    \small
    \bibliographystyle{ieeenat_fullname}
    \bibliography{main}

@String(CVPR= {IEEE Conf. Comput. Vis. Pattern Recog.})

@String(ICCV= {Int. Conf. Comput. Vis.})

@String(ECCV= {Eur. Conf. Comput. Vis.})

@String(BMVC= {Brit. Mach. Vis. Conf.})

@String(ICLR = {Int. Conf. Learn. Represent.})

@String(AAAI = {AAAI})

@String(CVPR  = {CVPR})

@String(ICCV  = {ICCV})

@String(ECCV  = {ECCV})

@String(BMVC  =	{BMVC})

@String(ICLR  = {ICLR})

@InProceedings{afreeca_dalessandro,
author="D'Alessandro, Adriano
and Mahdavi-Amiri, Ali
and Hamarneh, Ghassan",
editor="Leonardis, Ale{\v{s}}
and Ricci, Elisa
and Roth, Stefan
and Russakovsky, Olga
and Sattler, Torsten
and Varol, G{\"u}l",
title="AFreeCA: Annotation-Free Counting for All",
booktitle="Computer Vision -- ECCV 2024",
year="2025",
publisher="Springer Nature Switzerland",
address="Cham",
pages="75--91",
isbn="978-3-031-73235-5"
}

@article{xie2024described,
  title={Described object detection: Liberating object detection with flexible expressions},
  author={Xie, Chi and Zhang, Zhao and Wu, Yixuan and Zhu, Feng and Zhao, Rui and Liang, Shuang},
  journal={Advances in Neural Information Processing Systems},
  volume={36},
  year={2024}
}

@inproceedings{
podell2024sdxl,
title={{SDXL}: Improving Latent Diffusion Models for High-Resolution Image Synthesis},
author={Dustin Podell and Zion English and Kyle Lacey and Andreas Blattmann and Tim Dockhorn and Jonas M{\"u}ller and Joe Penna and Robin Rombach},
booktitle={The Twelfth International Conference on Learning Representations},
year={2024},
url={https://openreview.net/forum?id=di52zR8xgf}
}

@InProceedings{fake2real,
author="Qraitem, Maan
and Saenko, Kate
and Plummer, Bryan A.",
editor="Leonardis, Ale{\v{s}}
and Ricci, Elisa
and Roth, Stefan
and Russakovsky, Olga
and Sattler, Torsten
and Varol, G{\"u}l",
title="From Fake to Real: Pretraining on Balanced Synthetic Images to Prevent Spurious Correlations in Image Recognition",
booktitle="Computer Vision -- ECCV 2024",
year="2025",
publisher="Springer Nature Switzerland",
address="Cham",
pages="230--246",
isbn="978-3-031-73636-0"
}

@inproceedings{
he2023is,
title={Is Synthetic Data from Generative Models Ready for Image Recognition?},
author={Ruifei He and Shuyang Sun and Xin Yu and Chuhui Xue and Wenqing Zhang and Philip Torr and Song Bai and XIAOJUAN QI},
booktitle={The Eleventh International Conference on Learning Representations (ICLR)},
year={2023},
url={https://openreview.net/forum?id=nUmCcZ5RKF}
}

@techreport{WahCUB_200_2011,
	Title = {Caltech-UCSD Birds 200},
	Author = {Wah, C. and Branson, S. and Welinder, P. and Perona, P. and Belongie, S.},
	Year = {2011},
	Institution = {California Institute of Technology},
	Number = {CNS-TR-2011-001}
}

@inproceedings{nilsback2008automated,
  title={Automated flower classification over a large number of classes},
  author={Nilsback, Maria-Elena and Zisserman, Andrew},
  booktitle={2008 Sixth Indian conference on computer vision, graphics \& image processing},
  pages={722--729},
  year={2008},
  organization={IEEE}
}

@inproceedings{liu2023gres,
  title={Gres: Generalized referring expression segmentation},
  author={Liu, Chang and Ding, Henghui and Jiang, Xudong},
  booktitle={Proceedings of the IEEE/CVF conference on computer vision and pattern recognition},
  pages={23592--23601},
  year={2023}
}

@inproceedings{m_Ranjane_talCVPR21,
 author = {Viresh Ranjan and Udbhav Sharma and Thu Nguyen and Minh Hoai},
 title = {Learning To Count Everything},
 year = {2021},
 booktitle = {Proceedings of the {IEEE/CVF} Conference on Computer Vision and Pattern Recognition (CVPR)},
}

@InProceedings{va_count_zhu_2024,
    author="Zhu, Huilin
    and Yuan, Jingling
    and Yang, Zhengwei
    and Guo, Yu
    and Wang, Zheng
    and Zhong, Xian
    and He, Shengfeng",
    editor="Leonardis, Ale{\v{s}}
    and Ricci, Elisa
    and Roth, Stefan
    and Russakovsky, Olga
    and Sattler, Torsten
    and Varol, G{\"u}l",
    title="Zero-Shot Object Counting with Good Exemplars",
    booktitle="Computer Vision -- ECCV 2024",
    year="2025",
    publisher="Springer Nature Switzerland",
    address="Cham",
    pages="368--385",
    isbn="978-3-031-72652-1"
}

@inproceedings{liu2022countr,
  author = {Chang, Liu and Yujie, Zhong and Andrew, Zisserman and Weidi, Xie},
  title = {CounTR: Transformer-based Generalised Visual Counting},
  booktitle={British Machine Vision Conference (BMVC)},
  year = {2022}
}

@InProceedings{Pelhan_2024_CVPR,
    author    = {Pelhan, Jer and Luke\v{z}i\v{c}, Alan and Zavrtanik, Vitjan and Kristan, Matej},
    title     = {DAVE - A Detect-and-Verify Paradigm for Low-Shot Counting},
    booktitle = {Proceedings of the IEEE/CVF Conference on Computer Vision and Pattern Recognition (CVPR)},
    month     = {June},
    year      = {2024},
    pages     = {23293-23302}
}

@misc{Dukic_2023_ICCV,
      title={A Low-Shot Object Counting Network With Iterative Prototype Adaptation}, 
      author={Nikola Djukic and Alan Lukezic and Vitjan Zavrtanik and Matej Kristan},
      year={2023},
      eprint={2211.08217},
      archivePrefix={arXiv},
      primaryClass={cs.CV},
      url={https://arxiv.org/abs/2211.08217}, 
}

@inproceedings{countingdetr2022,
title     = {{Few-shot Object Counting and Detection}},
author    = {Thanh Nguyen, Chau Pham, Khoi Nguyen and Minh Hoai},
booktitle = {Proceedings of the European Conference on Computer Vision 2022},
year      = {2022}
}

@InProceedings{Dai_2024_CVPR,
    author    = {Dai, Siyang and Liu, Jun and Cheung, Ngai-Man},
    title     = {Referring Expression Counting},
    booktitle = {Proceedings of the IEEE/CVF Conference on Computer Vision and Pattern Recognition (CVPR)},
    month     = {June},
    year      = {2024},
    pages     = {16985-16995}
}

@inproceedings{zhizhong2024point,
  title={Point, Segment and Count: A Generalized Framework for Object Counting},
  author={Zhizhong, Huang and Mingliang, Dai and Yi, Zhang and Junping, Zhang and Hongming, Shan},
  booktitle={CVPR},
  year={2024}
}

@inproceedings{xu2023zero,
  title={Zero-shot object counting},
  author={Xu, Jingyi and Le, Hieu and Nguyen, Vu and Ranjan, Viresh and Samaras, Dimitris},
  booktitle={Proceedings of the IEEE/CVF Conference on Computer Vision and Pattern Recognition},
  pages={15548--15557},
  year={2023}
}

@inproceedings{shipard2023diversity,
  title={Diversity is definitely needed: Improving model-agnostic zero-shot classification via stable diffusion},
  author={Shipard, Jordan and Wiliem, Arnold and Thanh, Kien Nguyen and Xiang, Wei and Fookes, Clinton},
  booktitle={Proceedings of the IEEE/CVF Conference on Computer Vision and Pattern Recognition},
  pages={769--778},
  year={2023}
}

@INPROCEEDINGS{dalessandro10229913,
  author={D’ Alessandro, Adriano C. and Mahdavi-Amiri, Ali and Hamarneh, Ghassan},
  booktitle={2023 20th Conference on Robots and Vision (CRV)}, 
  title={Learning-to-Count by Learning-to-Rank}, 
  year={2023},
  volume={},
  number={},
  pages={105-112},
  doi={10.1109/CRV60082.2023.00021}}

@InProceedings{Liang_2023_CVPR,
    author    = {Liang, Dingkang and Xie, Jiahao and Zou, Zhikang and Ye, Xiaoqing and Xu, Wei and Bai, Xiang},
    title     = {CrowdCLIP: Unsupervised Crowd Counting via Vision-Language Model},
    booktitle = {Proceedings of the IEEE/CVF Conference on Computer Vision and Pattern Recognition (CVPR)},
    month     = {June},
    year      = {2023},
    pages     = {2893-2903}
}

@inproceedings{sam2019almost,
  title={Almost unsupervised learning for dense crowd counting},
  author={Sam, Deepak Babu and Sajjan, Neeraj N and Maurya, Himanshu and Babu, R Venkatesh},
  booktitle={Proceedings of the AAAI Conference on Artificial Intelligence},
  number={01},
  pages={8868--8875},
  year={2019}
}

@article{minderer2024scaling,
  title={Scaling open-vocabulary object detection},
  author={Minderer, Matthias and Gritsenko, Alexey and Houlsby, Neil},
  journal={Advances in Neural Information Processing Systems},
  volume={36},
  year={2024}
}

@inproceedings{
loshchilov2018decoupled,
title={Decoupled Weight Decay Regularization},
author={Ilya Loshchilov and Frank Hutter},
booktitle={International Conference on Learning Representations},
year={2019},
url={https://openreview.net/forum?id=Bkg6RiCqY7},
}

@inproceedings{avrahami2023bas,
  author = {Avrahami, Omri and Aberman, Kfir and Fried, Ohad and Cohen-Or, Daniel and Lischinski, Dani},
  title = {Break-A-Scene: Extracting Multiple Concepts from a Single Image},
  year = {2023},
  isbn = {9798400703157},
  publisher = {Association for Computing Machinery},
  address = {New York, NY, USA},
  url = {https://doi.org/10.1145/3610548.3618154},
  doi = {10.1145/3610548.3618154},        
  booktitle = {SIGGRAPH Asia 2023 Conference Papers},
  articleno = {96},
  numpages = {12},
  location = {, Sydney, NSW, Australia, },
  series = {SA '23}
}

@inproceedings{liang2023open,
  title={Open-vocabulary semantic segmentation with mask-adapted CLIP},
  author={Liang, F. and Wu, B. and Dai, X. and Li, K. and Zhao, Y. and Zhang, H. and Zhang, P. and Vajda, P. and Marculescu, D.},
  booktitle={CVPR},
  year={2023}
}

@InProceedings{AminiNaieni24,
  author = "Amini-Naieni, N. and Han, T. and Zisserman, A.",
  title = "CountGD: Multi-Modal Open-World Counting",
  booktitle = "Advances in Neural Information Processing Systems (NeurIPS)",
  year = "2024",
}

@inproceedings{chen2025single,
  title={Single Domain Generalization for Few-Shot Counting via Universal Representation Matching},
  author={Chen, Xianing and Huo, Si and Jiang, Borui and Hu, Hailin and Chen, Xinghao},
  booktitle={Proceedings of the Computer Vision and Pattern Recognition Conference},
  pages={4639--4649},
  year={2025}
}

@inproceedings{jia2022vpt,
  title={Visual Prompt Tuning},
  author={Jia, Menglin and Tang, Luming and Chen, Bor-Chun and Cardie, Claire and Belongie, Serge and Hariharan, Bharath and Lim, Ser-Nam},
  booktitle={European Conference on Computer Vision (ECCV)},
  year={2022}
}

@inproceedings{lester-etal-2021-power,
    title = "The Power of Scale for Parameter-Efficient Prompt Tuning",
    author = "Lester, Brian  and
      Al-Rfou, Rami  and
      Constant, Noah",
    editor = "Moens, Marie-Francine  and
      Huang, Xuanjing  and
      Specia, Lucia  and
      Yih, Scott Wen-tau",
    booktitle = "Proceedings of the 2021 Conference on Empirical Methods in Natural Language Processing",
    month = nov,
    year = "2021",
    address = "Online and Punta Cana, Dominican Republic",
    publisher = "Association for Computational Linguistics",
    doi = "10.18653/v1/2021.emnlp-main.243",
    pages = "3045--3059",
}

@inproceedings{zhou2022cocoop,
    title={Conditional Prompt Learning for Vision-Language Models},
    author={Zhou, Kaiyang and Yang, Jingkang and Loy, Chen Change and Liu, Ziwei},
    booktitle={IEEE/CVF Conference on Computer Vision and Pattern Recognition (CVPR)},
    year={2022}
}

@inproceedings{bianchi2024devil,
  title={The devil is in the fine-grained details: Evaluating open-vocabulary object detectors for fine-grained understanding},
  author={Bianchi, Lorenzo and Carrara, Fabio and Messina, Nicola and Gennaro, Claudio and Falchi, Fabrizio},
  booktitle={Proceedings of the IEEE/CVF Conference on Computer Vision and Pattern Recognition},
  pages={22520--22529},
  year={2024}
}

@InProceedings{lueddecke22_cvpr,
    author    = {L\"uddecke, Timo and Ecker, Alexander},
    title     = {Image Segmentation Using Text and Image Prompts},
    booktitle = {Proceedings of the IEEE/CVF Conference on Computer Vision and Pattern Recognition (CVPR)},
    month     = {June},
    year      = {2022},
    pages     = {7086-7096}
}

@misc{labs2025flux1kontextflowmatching,
      title={FLUX.1 Kontext: Flow Matching for In-Context Image Generation and Editing in Latent Space},
      author={Black Forest Labs and Stephen Batifol and Andreas Blattmann and Frederic Boesel and Saksham Consul and Cyril Diagne and Tim Dockhorn and Jack English and Zion English and Patrick Esser and Sumith Kulal and Kyle Lacey and Yam Levi and Cheng Li and Dominik Lorenz and Jonas Müller and Dustin Podell and Robin Rombach and Harry Saini and Axel Sauer and Luke Smith},
      year={2025},
      eprint={2506.15742},
      archivePrefix={arXiv},
      primaryClass={cs.GR},
      url={https://arxiv.org/abs/2506.15742},
}

@misc{esser2024scaling,
      title={Scaling Rectified Flow Transformers for High-Resolution Image Synthesis}, 
      author={Patrick Esser and Sumith Kulal and Andreas Blattmann and Rahim Entezari and Jonas Müller and Harry Saini and Yam Levi and Dominik Lorenz and Axel Sauer and Frederic Boesel and Dustin Podell and Tim Dockhorn and Zion English and Kyle Lacey and Alex Goodwin and Yannik Marek and Robin Rombach},
      year={2024},
      eprint={2403.03206},
      archivePrefix={arXiv},
      primaryClass={cs.CV},
      url={https://arxiv.org/abs/2403.03206}, 
}

@inproceedings{
hertz2023prompttoprompt,
title={Prompt-to-Prompt Image Editing with Cross-Attention Control},
author={Amir Hertz and Ron Mokady and Jay Tenenbaum and Kfir Aberman and Yael Pritch and Daniel Cohen-or},
booktitle={The Eleventh International Conference on Learning Representations },
year={2023}
}

@article{hobley2023abc,
        title={ABC Easy as 123: A Blind Counter for Exemplar-Free Multi-Class Class-agnostic Counting},
        author={Hobley, Michael and Prisacariu, Victor},
        journal={Proceedings of the European Conference on Computer Vision},
        year={2024}}

@InProceedings{Kirillov_2023_ICCV,
    author    = {Kirillov, Alexander and Mintun, Eric and Ravi, Nikhila and Mao, Hanzi and Rolland, Chloe and Gustafson, Laura and Xiao, Tete and Whitehead, Spencer and Berg, Alexander C. and Lo, Wan-Yen and Dollar, Piotr and Girshick, Ross},
    title     = {Segment Anything},
    booktitle = {Proceedings of the IEEE/CVF International Conference on Computer Vision (ICCV)},
    month     = {October},
    year      = {2023},
    pages     = {4015-4026}
}

@article{ravi2024sam2,
  title={SAM 2: Segment Anything in Images and Videos},
  author={Ravi, Nikhila and Gabeur, Valentin and Hu, Yuan-Ting and Hu, Ronghang and Ryali, Chaitanya and Ma, Tengyu and Khedr, Haitham and R{\"a}dle, Roman and Rolland, Chloe and Gustafson, Laura and Mintun, Eric and Pan, Junting and Alwala, Kalyan Vasudev and Carion, Nicolas and Wu, Chao-Yuan and Girshick, Ross and Doll{\'a}r, Piotr and Feichtenhofer, Christoph},
  journal={arXiv preprint arXiv:2408.00714},
  url={https://arxiv.org/abs/2408.00714},
  year={2024}
}

@article{liu2023grounding,
  title={Grounding dino: Marrying dino with grounded pre-training for open-set object detection},
  author={Liu, Shilong and Zeng, Zhaoyang and Ren, Tianhe and Li, Feng and Zhang, Hao and Yang, Jie and Li, Chunyuan and Yang, Jianwei and Su, Hang and Zhu, Jun and others},
    journal="Computer Vision -- ECCV 2024",
    year="2024",
    publisher="Springer Nature Switzerland",
    address="Cham"
}

@inproceedings{
keskar2017on,
title={On Large-Batch Training for Deep Learning: Generalization Gap and Sharp Minima},
author={Nitish Shirish Keskar and Dheevatsa Mudigere and Jorge Nocedal and Mikhail Smelyanskiy and Ping Tak Peter Tang},
booktitle={International Conference on Learning Representations},
year={2017},
url={https://openreview.net/forum?id=H1oyRlYgg}
}

@inproceedings{yun2019cutmix,
  title={Cutmix: Regularization strategy to train strong classifiers with localizable features},
  author={Yun, Sangdoo and Han, Dongyoon and Oh, Seong Joon and Chun, Sanghyuk and Choe, Junsuk and Yoo, Youngjoon},
  booktitle={Proceedings of the IEEE/CVF international conference on computer vision},
  pages={6023--6032},
  year={2019}
}
}


\end{document}